\documentclass[10pt,twocolumn,letterpaper]{article}

\usepackage{cvpr}
\usepackage{times}
\usepackage{graphicx}
\usepackage{amsmath}
\usepackage{amssymb}
\usepackage{epstopdf}
\graphicspath{{eps/}{jpeg/}}

\usepackage[breaklinks=true,bookmarks=false]{hyperref}

\cvprfinalcopy 

\def\cvprPaperID{1} 

\ifcvprfinal\fi
\begin{document}

\title{Condensation-Net: Memory-Efficient Network Architecture with\\ Cross-Channel Pooling Layers and Virtual Feature Maps}
\author{Tse-Wei Chen$^*$,
~Motoki Yoshinaga$^*$,
~Hongxing Gao$^\dag$,
~Wei Tao$^\dag$,
~Dongchao Wen$^\dag$,\\
~Junjie Liu$^\dag$,
~Kinya Osa$^*$,
~and Masami Kato$^*$\\
{\tt\small twchen@ieee.org}\\
\and
$^*$Device Technology Development Headquarters, Canon Inc.,\\
30-2, Shimomaruko 3-chome, Ohta-ku, Tokyo 146-8501, Japan\\
$^\dag$Canon Information Technology (Beijing) Co., Ltd.,\\
12A Floor, Yingu Building, No.9 Beisihuanxi Road, Haidian, Beijing, China\\
}


\maketitle
\thispagestyle{empty}

\begin{abstract}
``Lightweight convolutional neural networks" is an important research topic in the field of embedded vision. To implement image recognition tasks on a resource-limited hardware platform, it is necessary to reduce the memory size and the computational cost. The contribution of this paper is stated as follows. First, we propose an algorithm to process a specific network architecture (Condensation-Net) without increasing the maximum memory storage for feature maps. The architecture for virtual feature maps saves 26.5\% of memory bandwidth by calculating the results of cross-channel pooling before storing the feature map into the memory. Second, we show that cross-channel pooling can improve the accuracy of object detection tasks, such as face detection, because it increases the number of filter weights. Compared with Tiny-YOLOv2, the improvement of accuracy is 2.0\% for quantized networks and 1.5\% for full-precision networks when the false-positive rate is 0.1. Last but not the least, the analysis results show that the overhead to support the cross-channel pooling with the proposed hardware architecture is negligible small. The extra memory cost to support Condensation-Net is 0.2\% of the total size, and the extra gate count is only 1.0\% of the total size.


\end{abstract}


\begin{figure*}[h]
\begin{center}
   \includegraphics[width=0.9\linewidth]{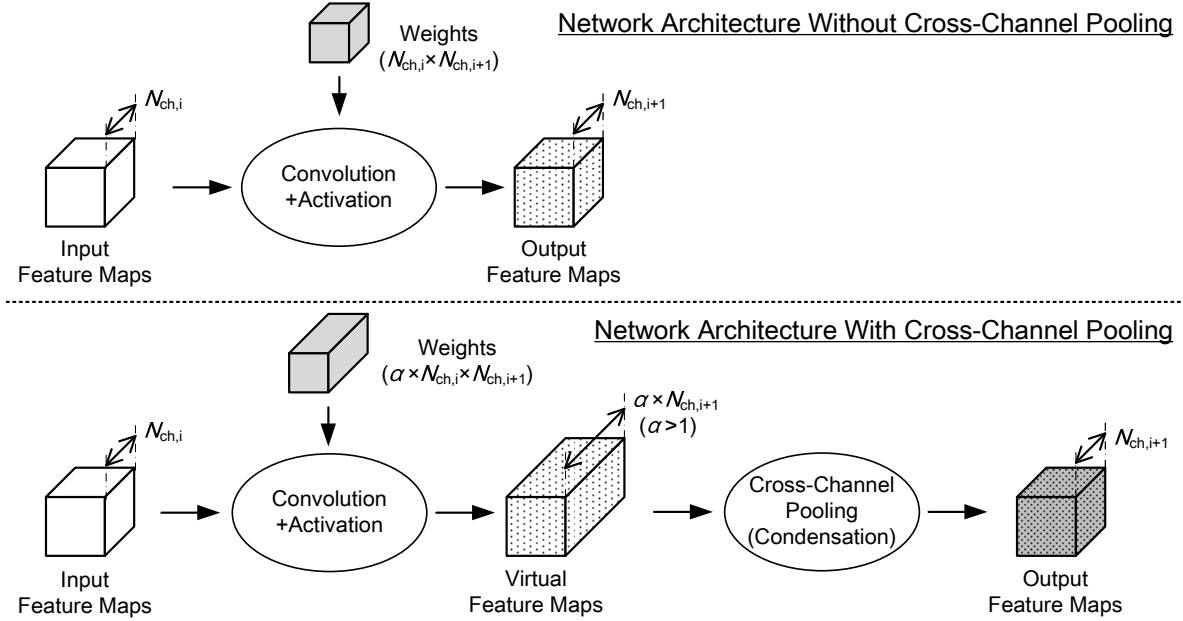}
\end{center}
   \caption{Concept of cross-channel pooling and virtual feature maps.}
\label{fig:concept}
\end{figure*}

\section{Introduction}
\label{sec:introduction}

Convolutional Neural Networks (CNNs) are widely used in image and video analysis applications, such as face alignment~\cite{Zhang16}, face recognition~\cite{Sun15}, object detection~\cite{Redmon17,Ren15}, scene segmentation~\cite{Chen17,Garcia17}, and so on. In order to implement these algorithms on mobile devices and embedded system platforms, it is necessary to reduce the memory size and the computational cost without decreasing the accuracy.

Many kinds of algorithms are proposed to handle lightweight networks on low-power devices. These algorithms can be classified into several categories. The first category is ``efficient network architecture design," where special operations are applied to reduce the size or the computationcal cost of a network. Howard, et al. propose a network architecture called MobileNets~\cite{Howard17,Sandler18}, where depth-wise separable convolutions are used to reduce the connections between layers. The second category is ``bit-width adjustment," where floating-point data are transferred to low-bit data to reduce the memory cost. Rastegari et al. propose a network architecture called XNOR-Net~\cite{Rastegari16}, where full-precision filter weights and full-precision feature maps are replaced with 1-bit filter weights and 1-bit feature maps, respectively, so that multiplication operations can be replaced with exclusive NOR (XNOR) operations. The third category is ``knowledge distillation," which is a common technique to increase the accuracy of small networks with extra information from other networks. Hinton et al. propose a technique to distill the knowledge from the teacher network into the student network~\cite{Hinton15} for training algorithms. When the size of student network is smaller than the teacher network, the technique can be used to increase the accuracy of small networks. In addition to the 3 categories, there are still many other kinds of algorithms for lightweight networks. These techniques can be applied to both hardware and software implementation.

For hardware implementation in embedded systems, it is important to achieve high performance and high recognition accuracy with compact network models. Boo et al. propose an architecture to compress the ternary weights by utilizing the structured sparsity~\cite{Boo17}, where a rule for look-up tables is applied to the training algorithm. Chen et al. propose a reconfigurable accelerator which contains a Run-Length Coding (RLC) module to compress the feature maps with consecutive zeros~\cite{YhChen17}. These techniques can be applied to different kinds of systems to reduce the network size without decreasing the accuracy, but it is difficult to find a systematic way to increase the accuracy of small networks.


In this paper, we propose a new approach, which is called Condensation-Net, to increase the accuracy of networks with limited hardware resources, including memory cost of feature maps and filter weights. The concept of the proposed method is shown in Figure~\ref{fig:concept}, which is separated into 2 parts. The upper part of Figure~\ref{fig:concept} shows the convolution process without cross-channel pooling, where the number of input feature maps and the number of output feature maps are $N_{\mathsf{ch},i}$ and $N_{\mathsf{ch},i+1}$, respectively. Both the input feature maps and the output feature maps are stored in the memory. The parameter $i$ represents the index of a convolution layer in the network. 

The lower part of Figure~\ref{fig:concept} shows the convolution process with cross-channel pooling operations, where the number of output feature maps is $\alpha N_{\mathsf{ch},i+1}$ and the number of output feature maps after cross-channel pooling is $N_{\mathsf{ch},i+1}$. Since $\alpha$ is larger than 1, we can use more filter weights to compute the output feature maps than the network architecture in the upper part of Figure~\ref{fig:concept}. After computing the output feature maps, cross-channel pooling operations are applied, and $\alpha N_{\mathsf{ch},i+1}$ output feature maps will be condensed into $N_{\mathsf{ch},i+1}$ output feature maps. It is not necessary to store all the output feature maps before cross-channel pooling because the result of cross-channel pooling can be computed sequentially with only a part of output feature maps. That is, we can increase the accuracy of the network by adding filter weights while keeping the number of stored feature maps the same. For embedded computing with limited hardware resources, it is an advantage to improve the accuracy of the network without increasing the memory storage of feature maps.

There are two main ideas in our approach, cross-channel pooling and virtual feature maps. Cross-channel pooling is a technique commonly used to combine the information of multiple feature maps (channels). Marcos, et al. use cross-channel pooling operations to preserve the rotation-invariant features for texture classification~\cite{Marcos16}. Laptev et al. apply an operator called ``Transform-Invariant pooling (TI-Pooling)" to the fully-connected layers~\cite{Laptev16}. Nguyen et al. propose a sparse temporal pooling network, where a video-level representation is generated via weighted temporal average pooling~\cite{Nguyen17}. In this paper, we use cross-channel pooling for ``condensation," which means to reduce the number of channels while preserving the information. The input feature maps of cross-channel pooling are called ``virtual feature maps," which are computed sequentially and NOT stored in the main memory. 

The paper is organized as follows. In Sec.~\ref{sec:algorithm}, the proposed networks architecture and the algorithm are introduced. The proposed hardware architecture is shown in Sec.~\ref{sec:architecture}. The experimental results are discussed in Sec.~\ref{sec:results}. The conclusions are given in Sec.~\ref{sec:conclusion}.

\section{Proposed Network: Condensation-Net}
\label{sec:algorithm}

\begin{figure*}[h]
\begin{center}
   \includegraphics[width=0.95\linewidth]{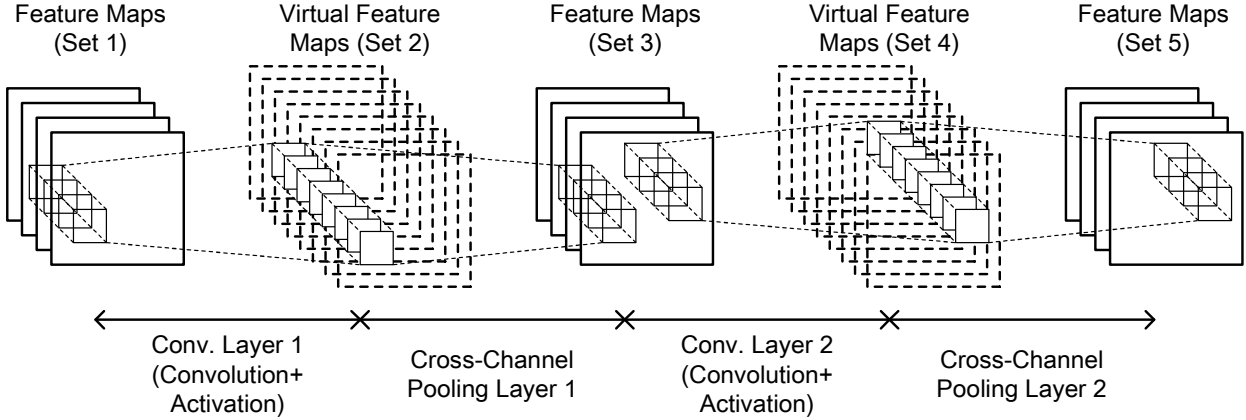}
\end{center}
   \caption{Example of the proposed network, Condensation-Net.}
\label{fig:network}
\end{figure*}

An example of the proposed network, Condensation-Net, is shown in Figure~\ref{fig:network}. There are 5 sets of feature maps, 2 convolution layers, and 2 cross-channel pooling layers in the network. The 1st set and the 3rd set of feature maps are stored in the memory, but the 2nd set and the 4th set of feature maps, which are the input of the 1st cross-channel pooling layer and the 2nd cross-channel pooling layer, respectively, are not stored in the memory. The 2nd set and the 4th set of feature maps, which are shown in the dotted lines, are also called ``virtual feature maps." The 2nd set of feature maps is the output of the 1st convolution layer and is calculated based on the 1st set of feature maps with convolution and activation functions. The 3rd set of feature maps is the output of the 1st cross-channel pooling layer and is calculated based on the 2nd set of feature maps with cross-channel pooling functions.


\begin{figure}[t]
\begin{center}
   \includegraphics[width=0.95\linewidth]{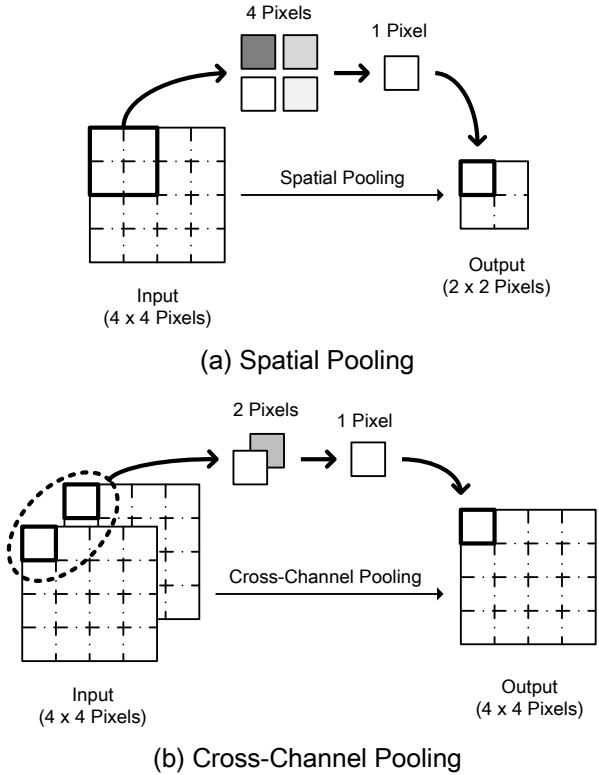}
\end{center}
   \caption{Illustration of spatial pooling and cross-channel pooling.}
\label{fig:pooling}
\end{figure}

\begin{figure*}[t]
\begin{center}
   \includegraphics[width=0.8\linewidth]{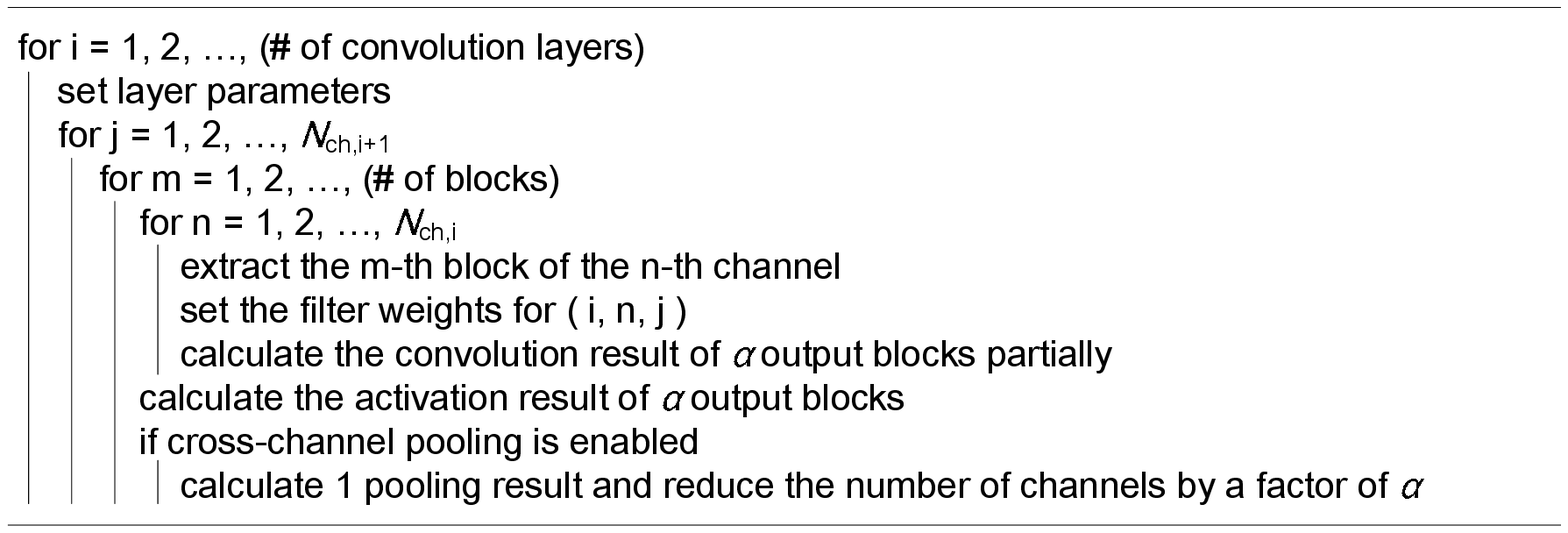}
\end{center}
   \caption{Proposed algorithm.}
\label{fig:flowchart}
\end{figure*}

\begin{figure*}[t]
\begin{center}
   \includegraphics[width=0.9\linewidth]{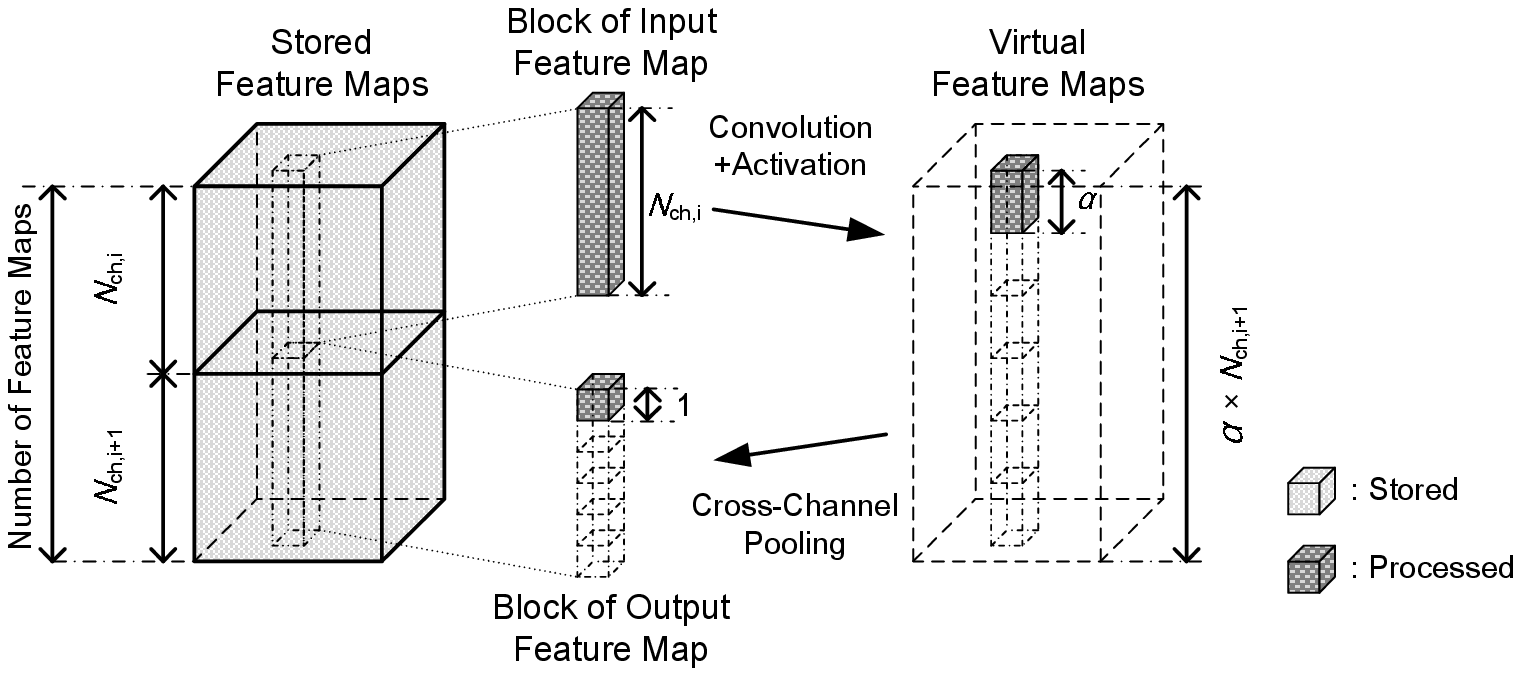}
\end{center}
   \caption{Relation between virtual feature maps and memory.}
\label{fig:relation}
\end{figure*}

\subsection{Cross-Channel Pooling}
\label{subsec:ccpooling}

Figure~\ref{fig:pooling} shows the comparison of spatial pooling operations and cross-channel pooling operations. An illustration of spatial pooling is shown in Figure~\ref{fig:pooling}(a), where the stride and the window size are both 2 pixels (2 $\times$ 2 pixels) in the spatial domain. The size of input feature map is reduced from 4 $\times$ 4 pixels to 2 $\times$ 2 pixels in the spatial domain. The number of feature maps does not change. An illustration of cross-channel pooling is shown in Figure~\ref{fig:pooling}(b), where the stride and the window size are both 2 pixels in the channel direction. In this example, the window size for cross-channel pooling has no connection with the spatial domain. The number of input feature maps is 2, and the number of output feature maps is reduced to 1. The resolution of feature maps does not change after cross-channel pooling. The operations of spatial pooling and cross-channel pooling are similar, but the numbers of filter weights required to compute the input feature maps for the 2 pooling algorithms can be different. The number of filter weights for a convolution layer with cross-channel pooling operations might be 2 times larger than the one with spatial pooling operations. The output pixel of cross-channel pooling can be computed based on the max operations, the average operations, the min operations, and so on.

Figure~\ref{fig:flowchart} shows the proposed algorithms. There are 4 levels in the nested loop structure. In the 1st level of loops, each convolution layer of the networks is processed sequentially, and the parameters of the corresponding layer are set. In the 2nd level of loops, each output channel of the layer is processed sequentially. In the 3rd level of loops, each block of the output channel is processed sequentially. In the 4th level of loops, each input channel is processed sequentially. The filter weights of the $i$-th convolution layer, the $j$-th output channel, the $n$-th input channel are set, and the convolution results of $\alpha$ output blocks are computed. If cross-channel pooling operations are enabled, the output block to the next layer is calculated and the number of feature maps becomes 1 since the output blocks are reduced by a factor of $\alpha$ in the channel direction. Otherwise, the output result to the next layer includes the convolution results of $\alpha$ blocks (channels) and no operations are required.

\begin{figure*}[t]
\begin{center}
   \includegraphics[width=0.9\linewidth]{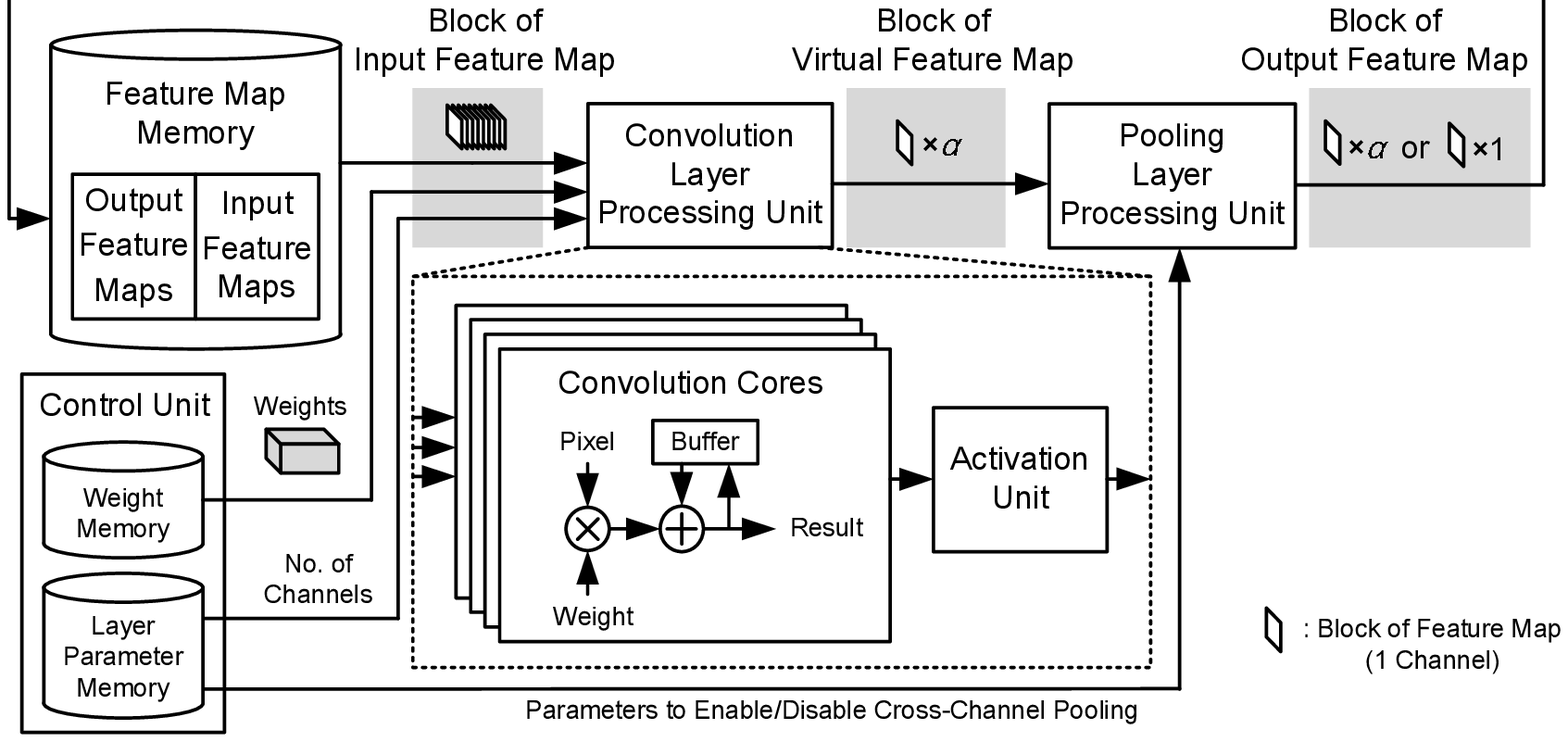}
\end{center}
   \caption{Overview of the proposed hardware architecture.}
\label{fig:architecture}
\end{figure*}

\subsection{Virtual Feature Maps}
\label{subsec:vitrual}

Figure~\ref{fig:relation} shows the relation between the ``virtual feature maps" and the feature map memory. There are $N_{\mathsf{ch},i}$ input feature maps (channels) in the upper part of the feature map memory and $N_{\mathsf{ch},i+1}$ output feature maps in the lower part of the feature map memory. A block is extracted from the input feature maps, and the result of convolution and activation, which is a part of virtual feature maps, is generated. There are $\alpha N_{\mathsf{ch},i+1}$ virtual feature maps in total. After cross-channel pooling, the number of output feature maps is reduced to $N_{\mathsf{ch},i+1}$, and a block of the output feature maps is generated. The same operation are repeated until all blocks are processed.

The memory can store $N_{\mathsf{ch},i}$ channels for the input and $N_{\mathsf{ch},i+1}$ channels for the output. If all the virtual feature maps are stored, we need an extra storage for $\alpha  N_{\mathsf{ch},i+1}$ channels. Since the result of cross-channel pooling can be calculated partially, it is not necessary to store all virtual feature maps. By employing the proposed algorithm, only the storage of 1 block is required to compute the result of cross-channel pooling. The results of cross-channel pooling are also called virtual feature maps since it is not necessary to store them in the physical memory.

\subsection{Activation Functions Based on Quantization}

The activation functions mentioned in the previous sections can be implemented by using Rectified Linear Unit (ReLU) functions, sigmoid functions, tangent functions, leaky ReLU functions, and so on. When the bit width of feature maps and filter weights are small, the activation functions can also be implemented with quantization functions~\cite{Zhou16,Cai17}. The zeros in the feature maps can be removed by using cross-channel pooling operations without losing much information especially for low-bit networks (e.g. 1 bit or 2 bits) since the pixels of feature maps contain lots of zeros.

\subsection{Training Algorithm}
\label{subsec:training}

The proposed network with cross-channel pooling can be trained by using the back-propagation algorithm. The following is an example of the training algorithm for cross-channel pooling with the max operations.

For a cross-channel pooling layer with a stride of 2 pixels and a window size of 2 pixels in the channel direction, the maximum value of 2 input pixels is computed in the forward-propagation process, and the corresponding channel which has the maximum values is recorded. Then, the gradient value is sent to the recorded channel of the cross-channel pooling layer in the back-propagation process. The steps of forward-propagation process and back-propagation process are repeated until the training results converge.


\section{Proposed Hardware Architecture}
\label{sec:architecture}

Figure~\ref{fig:architecture} shows the proposed hardware architecture. The feature map memory is used to store the input feature maps and the output feature maps, which are the feature maps in 2 consecutive layers in a network. The weight memory is used to store the filter weights, and the layer parameter memory is used to store layer parameters, which indicate the size of filter weights, the interconnections of input and output channels, the stride and the window size of pooling operations, and so on. The convolution results are computed in the convolution cores, and the activation results are computed in the activation unit. The convolution cores and the activation unit are included in the ``Convolution Layer Processing Unit (CLPU)," and the result of cross-channel pooling is computed in the ``Pooling Layer Processing Unit (PLPU)." The result of cross-channel pooling from the PLPU is a set of blocks of feature maps, which are store in the feature map memory as a part of feature maps.

When the number of input channel is $N_{\mathsf{ch},i}$, the convolution unit needs to process $N_{\mathsf{ch},i}$ blocks of input feature maps before generating 1 block of virtual feature map. The convolution unit contains $M$ convolution cores, which are able to process $M$ convolution operations in parallel. To accelerate the computations, it is also feasible to generate multiple virtual feature maps for different channels by using multiple CLPUs. The value of $M$ is set according to the size of network, the requirement of computational speed, and the target size of the hardware. Since the activation unit and the convolution cores work in a pipeline manner, the activation result of the block can be computed when the result of cross-channel pooling is generated. The parameters of PLPU are set according to the layer parameters. When the cross-channel pooling is not enabled, the virtual feature maps are the same as the output feature maps.

\section{Experimental Results}
\label{sec:results}

The experiments results contain 2 parts. The first part is the comparison of accuracy of face detection. The second part is the analysis of the memory cost of the proposed hardware architecture.


\subsection{Comparison of Accuracy}
\label{subsec:comparison}

We evaluate the 2 networks shown in Table~\ref{tab:network}. The first one is Tiny-YOLOv2, which is a compact network for object detection~\cite{Redmon17, Wai18}. The second one is the proposed Condensation-Net, where the parameters are extended from Tiny-YOLOv2. In Condensation-Net, the 1st -- the 4th convolution layers are replaced by the combinations of convolution layers and cross-channel pooling layers shown in Figure~\ref{fig:concept}. The reason to choose the 1st -- the 4th convolution layers is that the memory cost of filter weights in these layers is relatively small compared to the total cost. In the 1st convolution layer, the filter size of Condensation-Net is $16 \alpha \times 3 \times 3$, which means that the number of output channels, $N_{\mathsf{ch},1} $, is $16 \alpha$, and both the width and the height are 3. The number of input channels is 3 since there are 3 channels (RGB) in the input image. For Condensation-net, the value of $\alpha$ is set to 2 or 4, and Tiny-YOLOv2 can be regarded as a special case of Condensation-Net when $\alpha = 1$. The cross-channel pooling layers can be added to different layers according to the requirements of applications.

\begin{figure}[t]
\begin{center}
   \includegraphics[width=1.0\linewidth]{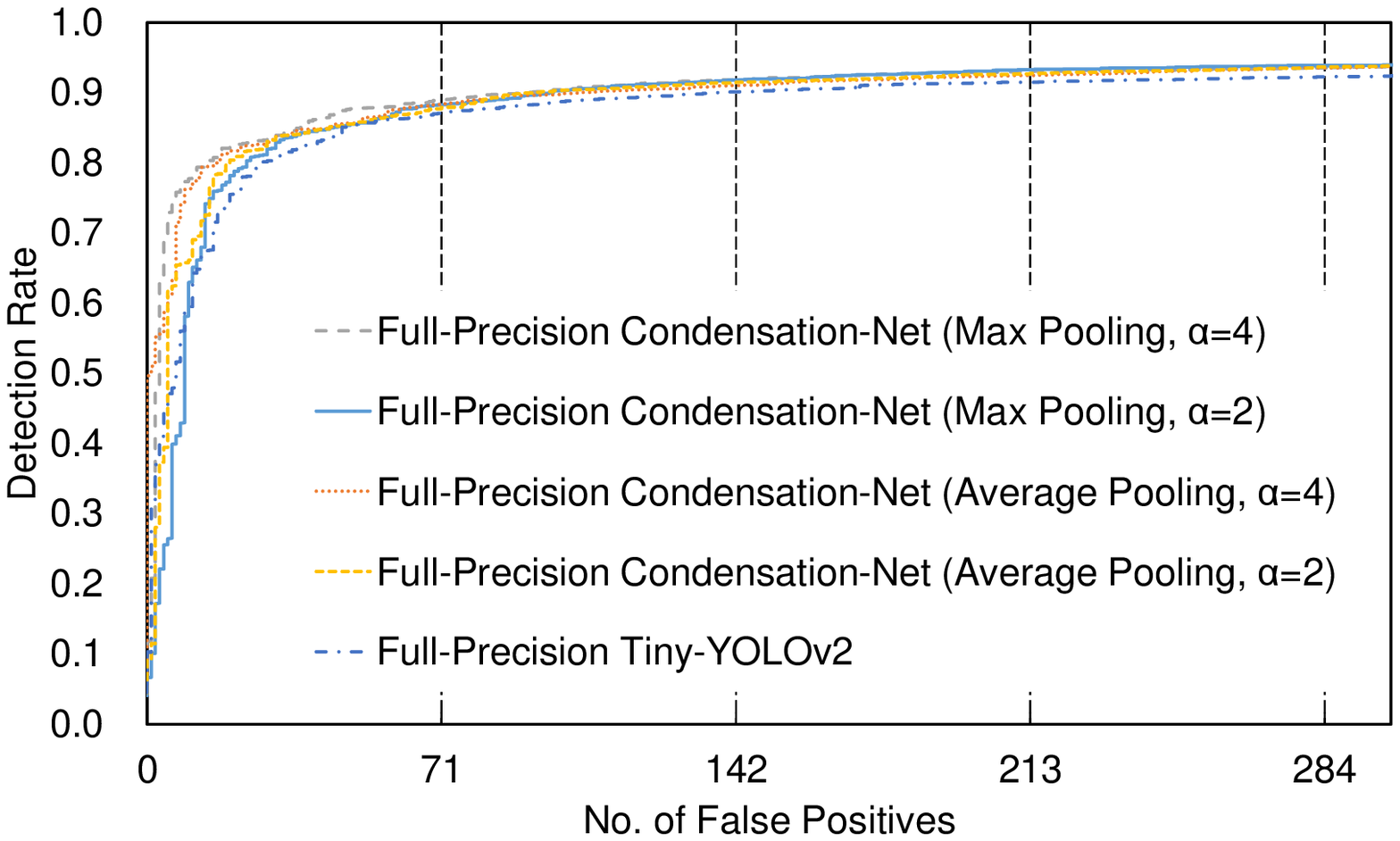}\\
   (a)\\
   \includegraphics[width=1.0\linewidth]{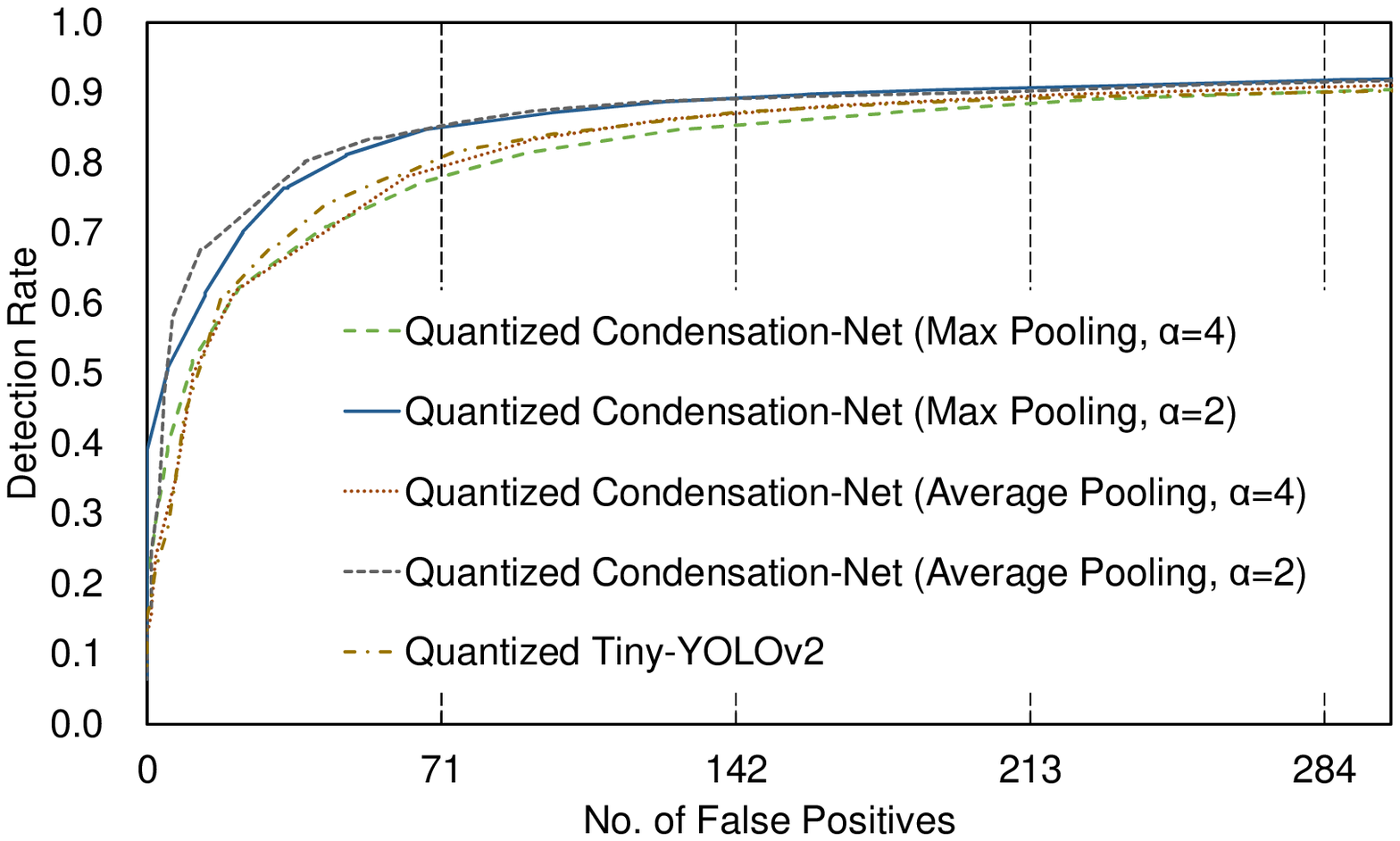}\\
   (b)\\
\end{center}
   \caption{Comparison of ROC curves for (a) full-precision networks and (b) quantized networks.}
\label{fig:roc}
\end{figure}

The images in the training set of Wider Face~\cite{Yang16} are used to train the 2 networks, Tiny-YOLOv2 and Condensation-Net. We compare the accuracy of the networks on the Face Detection Data Set and Benchmark (FDDB)~\cite{Jain10}, which contains 5,171 faces in 2,845 test images. The ROC curves of the networks are shown in Figure~\ref{fig:roc}, which shows that Condensation-Net achieves better detection rate than Tiny-YOLOv2 especially when the number of false positive is small. To further evaluate the performance of the face detector, we compare the detection rate (true-positive rate) when the false-positive rate is 0.1 (1 false positive / 10 test images), which corresponds to $\lfloor 2,845 \times 0.1 \rfloor = 284$ faces in 2,845 test images. The comparison results are shown in Table~\ref{tab:accuracy}. 

As mentioned in Sec.~\ref{sec:introduction}, one method to reduce the memory size is quantization. We quantize the network using the Half-Wave Gaussian Quantization (HWGQ) algorithm, where full-precision filter weights and full-precision feature maps are replaced with 1-bit filter weights and 2-bit feature maps, respectively~\cite{Cai17}. For quantized networks and full-precision networks, Condensation-Net achieves higher accuracy than Tiny-YOLOv2 because Condensation-Net contains more filter weights than Tiny-YOLOv2. It means that the proposed cross-channel pooling layers can be applied to either quantized networks or full-precision networks to increase the accuracy. However, when $\alpha$ is increased from 2 to 4, the number of filter weights in the 1st -- the 4th convolution layer doubles, but the accuracy decreases. The reason can be that some information contained in the parameters, which is essential to increasing the accuracy of face detection, is removed by cross-channel pooling. Similar to spatial pooling operations, where the window size and the stride are set to $2 \times 2$ pixels for many applications, the parameter $\alpha = 2$ is found to be the optimal value for the proposed cross-channel pooling on face detection tasks according to the experimental result.

In the spatial pooling layers, we re-use the parameters of Tiny-YOLOv2 and employ the max operations to compute the result of spatial pooling. However, to compare the accuracy, we employ the max operations and the average operations to compute the result of cross-channel pooling in the 1st -- the 4th convolution layers. The accuracy of these two kinds of operations is similar, but the max operations achieve slightly better performance than the average operations except for the full-precision network with $\alpha = 2$. According to the experimental results, we employ the max operations for cross-channel pooling on face detection tasks.

\begin{table}
\begin{center}
\begin{tabular}{l|c|c}
\hline
  & \multicolumn{2}{c}{Filter Size / Stride$^{*}$}   \\
  \cline{2-3}
  & Tiny-YOLOv2 & Condensation-Net  \\
  &    ~\cite{Redmon17, Wai18}  & (Our Work)  \\
\hline\hline
Conv. Layer 1    & $16 \times 3 \times 3$    & $16 \alpha \times 3 \times 3$ / $\alpha$\\
\hline
Pooling Layer 1  & \multicolumn{2}{c}{$2 \times 2$ / $2$}\\          
\hline
Conv. Layer 2    & $32 \times 3 \times 3$    & $32 \alpha \times 3 \times 3$ / $\alpha$\\
\hline
Pooling Layer 2  & \multicolumn{2}{c}{$2 \times 2$ / $2$}\\          
\hline
Conv. Layer 3    & $64 \times 3 \times 3$    & $64 \alpha \times 3 \times 3$ / $\alpha$\\
\hline
Pooling Layer 3  & \multicolumn{2}{c}{$2 \times 2$ / $2$}\\          
\hline
Conv. Layer 4    & $128 \times 3 \times 3$   & $128 \alpha \times 3 \times 3$ / $\alpha$\\
\hline
Pooling Layer 4  & \multicolumn{2}{c}{$2 \times 2$ / $2$}\\          
\hline
Conv. Layer 5    & \multicolumn{2}{c}{$256 \times 3 \times 3$}\\          
\hline
Pooling Layer 5  & \multicolumn{2}{c}{$2 \times 2$ / $2$}\\          
\hline
Conv. Layer 6    & \multicolumn{2}{c}{$512 \times 3 \times 3$}\\          
\hline
Pooling Layer 6  & \multicolumn{2}{c}{$2 \times 2 $}\\          
\hline
Conv. Layer 7    & \multicolumn{2}{c}{$1024 \times 3 \times 3$}\\          
\hline
Conv. Layer 8    & \multicolumn{2}{c}{$1024 \times 3 \times 3$}\\          
\hline
Conv. Layer 9    & \multicolumn{2}{c}{$30 \times 1 \times 1$}\\          
\hline
\end{tabular}
\end{center}
{\small
$^{*}$The stride in the 1st -- the 4th convolution layers ($\alpha$) in Condensation-Net represents the stride in the channel direction, not in the spatial domain. The stride for the pooling layers represents the stride in the spatial domain, not in the channel direction.\\
}
\caption{Network Architecture and Filter Weights of Tiny-YOLOv2 and Condensation-Net}
\label{tab:network}
\end{table}

\begin{table}
\begin{center}
\begin{tabular}{l|ccc}
\hline
           & Quantized & Full-Precision \\
           & Network   & Network \\
\hline\hline
Tiny-YOLOv2                        &  89.87\% & 92.28\% \\
\hline
Condensation-Net\\
$\alpha = 2$, Max Pooling$^{1}$    &  91.82\% & 93.86\% \\
$\alpha = 2$, Avg. Pooling$^{2}$   &  91.13\% & 93.69\% \\
$\alpha = 4$, Max Pooling$^{1}$    &  90.25\% & 93.61\% \\
$\alpha = 4$, Avg. Pooling$^{2}$   &  90.74\% & 93.56\% \\
\hline
\end{tabular}
\end{center}
{\small
$^{1}$Cross-channel pooling with the max operations.\\
$^{2}$Cross-channel pooling with the average operations.\\
}
\caption{Accuracy of Face Detection (False-Positive Rate = 0.1)}
\label{tab:accuracy}
\end{table}

\subsection{Analysis of Memory Size}
\label{subsec:memory}

The comparison of the required memory size of the proposed hardware architecture for different networks is shown in Table~\ref{tab:size}. As shown in Figure~\ref{fig:architecture}, the proposed hardware architecture stores the data of 2 consecutive feature maps and all of the filter weights in the network. The maximum size of input images is $512 \times 512$ pixels $\times 8$ bits. For the convolution layers where $\alpha$ is set to 2 or 4, we only need to store $1/\alpha$ of feature maps after convolution operations because of cross-channel pooling. The size of the weight memory is 1,924 KB for Tiny-YOLOv2, and the sizes of the weight memory are 1,935 KB and 1,959 KB for Condensation-Net when $\alpha = 2$ and $\alpha = 4$, respectively. The memory sizes of the feature map memory for Tiny-YOLOv2 and Condensation-Net ($\alpha = 2$ or $\alpha = 4$) are both 4,096 KB. They are exactly the same since we do not have to store the virtual feature maps as mentioned in Sec.~\ref{subsec:vitrual}. The numbers of feature maps of Tiny-YOLOv2 and Condensation-Net are different after convolution operations, but they become the same after applying cross-channel pooling to Condensation-Net. The memory sizes of the weight memory for Tiny-YOLOv2 and Condensation-Net are slightly different because Condensation-Net requires more filter weights than Tiny-YOLOv2 to compute the results of convolution. However, as shown in Table~\ref{tab:network}, since the numbers of channels in the 1st -- the 4th convolution layers are small, the difference of filter weights between the 2 networks is also small.

As shown in Table~\ref{tab:accuracy} and Table~\ref{tab:size}, the total size of the weight memory and the feature map memory for Tiny-YOLOv2 is 6,020 KB, and the detection rate of face detection is 89.87\%. By using the proposed techniques, we can increase the detection rate by adding memory storage with negligible costs. The total size of the weight memory and the feature map memory for Condensation-Net ($\alpha = 2$) is 6,031 KB, but the detection rate of face detection is 91.82\%, which is higher than Tiny-YOLOv2. The overhead of the proposed hardware architecture is 11 KB, which is only 0.2\% of the total memory size.

\begin{table}
\begin{center}
\begin{tabular}{l|ccc}
\hline
           & \multicolumn{2}{c}{Memory Size}\\  
           & Total & (Weight / Feature Map)\\
\hline\hline
Tiny-YOLOv2                        &  6,020 KB & (1,924 KB / 4,096 KB) \\
\hline
Condensation-Net\\
$\alpha = 2$                     &  6,031 KB & (1,935 KB / 4,096 KB) \\
$\alpha = 4$                     &  6,055 KB & (1,959 KB / 4,096 KB) \\
\hline
\end{tabular}
\end{center}
\caption{Memory Size of Proposed Hardware Architecture}
\label{tab:size}
\end{table}

\begin{figure}[t]
\begin{center}
   \includegraphics[width=0.95\linewidth]{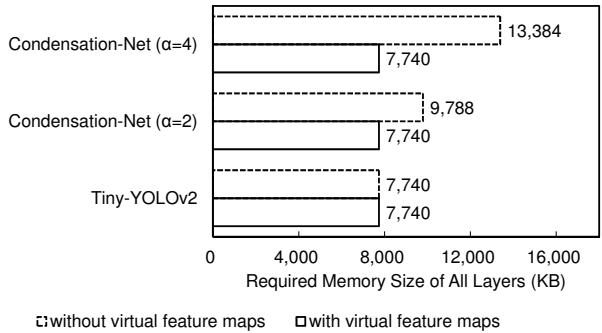}
\end{center}
   \caption{Comparison of the required memory size of all layers.}
\label{fig:bandwidth}
\end{figure}

Figure~\ref{fig:bandwidth} shows the comparison of the required memory size of all layers. In our hardware architecture, we only need to store the feature map of 2 consecutive layers, but the memory access of feature maps in all layers is inevitable. When $\alpha = 2$, the required memory size of Condensation-Net is 9,788 KB, but we can reduce the memory access to 7,740 KB with the virtual feature maps. It means that we can save 26.5\% of memory bandwidth. The larger the feature map, the more memory access can be saved.

The specifications of the proposed hardware architecture, which is designed to process the quantized Condensation-Net ($\alpha = 2$), are shown in Table~\ref{tab:spec}. We use the 28-nm CMOS technology library for the experiments. The performance is 320 Multiply-Accumulate operations (MACs) per clock cycle for 2-bit feature maps and 1-bit filter weights, which means that the value of $M$ is set to 320. As shown in Figure~\ref{fig:architecture}, since the cross-channel pooling operations can be enabled or disabled in different layers, the hardware architecture can also process Tiny-YOLOv2, which includes no cross-channel pooling operations. The gate counts of the CLPU and the PLPU are 197K and 2K, respectively. The gate count of the PLPU is a small number compared to the gate count of the whole hardware architecture, 199K. It means that we can support cross-channel pooling functions by increasing only 1.0\% of the hardware resources. Compared with Tiny-YOLOv2, it takes 1.3 times of processing time to compute the inference result of Condensation-Net. Since the proposed hardware can support both Tiny-YOLOv2 and Condensation-Net, the target networks can be switched according to the target processing time.

\begin{table}
\begin{center}
\begin{tabular}{lccc}
\hline
  Memory Size & Feature Map Memory: 4,096 KB   \\
              & Weight Memory: 1,935 KB   \\
\hline
  Gate Count & CLPU$^{1}$: 197K \\
  (NAND-Gates) & PLPU$^{2}$: 2K\\  
\hline
  Process &  28-nm CMOS Technology\\
\hline
  Frequency &  400 MHz\\  
\hline
  Input Image Size & $512 \times 512$ pixels\\
\hline
  Processing Speed & Tiny-YOLOv2$^{3}$: 95 ms\\
                   & Condensation-Net$^{3}$ ($\alpha = 2$): 124 ms\\
\hline
  Filter Size$^{4}$ & Maximum: $N_{\mathsf{ch},i} \times 7 \times 7$\\
\hline
  Performance &   320 MACs / clock cycle\\
\hline
\end{tabular}
\end{center}
{\small
$^{1}$CLPU stands for Convolution Layer Processing Unit.\\
$^{2}$PLPU stands for Pooling Layer Processing Unit.\\
$^{3}$Tiny-YOLOv2 and Condensation-Net are quantized networks.\\
$^{4}N_{\mathsf{ch},i}$ represents the number of output channels in the $i$-th convolution layer in the network.\\
}
\caption{Specifications of Proposed Hardware Architecture}
\label{tab:spec}
\end{table}

The comparison of hardware architectures for low-bit quantized CNNs is shown in Table \ref{tab:related}. It is difficult to compare the proposed hardware architecture with the related works~\cite{li17,zhao17} because they are implemented on FPGA platforms. Since the proposed architecture is synthesized with the cell-based design library, it generally achieves higher clock frequency than FPGAs. The proposed hardware architecture can achieve 7.72 GOPS / kLUT for 1-bit -- 2-bit operations. The performance density is higher than the related work~\cite{zhao17} without considering the cost of DSPs in FPGA, and the supported bit width is larger than the related work ~\cite{li17}.

\begin{table}
\begin{center}
\begin{tabular}{l|cccc}
\hline
               & Device          &  Clock & Bit        & Performance \\
               &                 &        & Width      & Density\\
               &                 &  (MHz) & (bits)      & (GOPS / kLUT)\\
\hline\hline
~\cite{li17}   & FPGA           & 90    & 1               & 22.40\\
~\cite{zhao17} & FPGA           & 143   & 1 -- 2          & 4.43 \\
Ours           & ASIC           & 400   & 1 -- 2          & 7.72$^{*}$\\
\hline
\end{tabular}
\end{center}
{\small
$^{*}$1 LUT = 6 two-input NAND gates.\\
}
\caption{Comparison of Hardware Architectures for Low-Bit Quantized CNNs}
\label{tab:related}
\end{table}



\section{Conclusions and Future Work}
\label{sec:conclusion}

In this paper, we propose a new network architecture, Condensation-Net, which combines cross-channel pooling with a specific processing order for virtual feature maps. The experimental results show that the proposed algorithm achieves higher accuracy than Tiny-YOLOv2 for face detection applications, where the dataset of FDDB are used for training and the Wider Face dataset are used for testing. The overhead of memory size to support the cross-channel pooling functions is only 0.2\%, and the extra gate count is only 2K gates. Besides, the architecture for virtual feature maps saves 26.5\% of memory bandwidth by calculating the results of cross-channel pooling before storing the feature map into the memory.

The proposed network can be implemented on different hardware platforms. For future work, we plan to apply the proposed method to different applications, including image segmentation and face recognition, and test the accuracy of specific image-recognition tasks. In addition to Tiny-YOLOv2, we will apply cross-channel pooling to other kinds of networks and analyze the tradeoff among the memory cost, the processing speed, and the accuracy.

{\small
\bibliographystyle{ieee_fullname}
\bibliography{IEEEfull,egbib}

\begin{thebibliography}{10}\itemsep=-1pt

\bibitem{Boo17}
Yoonho Boo and Wonyong Sung.
\newblock Structured sparse ternary weight coding of deep neural networks for
  efficient hardware implementations.
\newblock In {\em Proceedings of IEEE International Workshop on Signal
  Processing Systems (SiPS)}, Oct. 2017.

\bibitem{Cai17}
Zhaowei Cai, Xiaodong He, Jian Sun, and Nuno Vasconcelos.
\newblock Deep learning with low precision by half-wave {Gaussian}
  quantization, 2017.
\newblock arXiv:1702.00953.

\bibitem{Chen17}
Liang-Chieh Chen, George Papandreou, Iasonas Kokkinos, Kevin Murphy, and
  Alan~L. Yuille.
\newblock {DeepLab: Semantic} image segmentation with deep convolutional nets,
  {Atrous} convolution, and fully connected {CRF}s, 2017.
\newblock arXiv:1606.00915.

\bibitem{YhChen17}
Yu-Hsin Chen, Tushar Krishna, Joel~S. Emer, and Vivienne Sze.
\newblock {Eyeriss: An} energy-efficient reconfigurable accelerator for deep
  convolutional neural networks.
\newblock {\em IEEE Journal of Solid-State Circuits}, 52(1):127--138, 2017.

\bibitem{Garcia17}
Alberto Garcia-Garcia, Sergio Orts-Escolano, Sergiu Oprea, Victor
  Villena-Martinez, and Jose Garcia-Rodriguez.
\newblock A review on deep learning techniques applied to semantic
  segmentation, 2017.
\newblock arXiv:1704.06857.

\bibitem{Hinton15}
Geoffrey Hinton, Oriol Vinyals, and Jeff Dean.
\newblock Distilling the knowledge in a neural network, 2015.
\newblock arXiv:1503.02531.

\bibitem{Howard17}
Andrew~G. Howard, Menglong Zhu, Bo Chen, Dmitry Kalenichenko, Weijun Wang,
  Tobias Weyand, Marco Andreetto, and Hartwig Adam.
\newblock {MobileNets: Efficient} convolutional neural networks for mobile
  vision applications, 2017.
\newblock arXiv:1704.04861.

\bibitem{Jain10}
Vidit Jain and Erik Learned-Miller.
\newblock {FDDB: A} benchmark for face detection in unconstrained settings.
\newblock Technical Report UM-CS-2010-009, University of Massachusetts,
  Amherst, 2010.

\bibitem{Laptev16}
Dmitry Laptev, Nikolay Savinov, Joachim~M. Buhmann, and Marc Pollefeys.
\newblock {TI-POOLING: Transformation-invariant} pooling for feature learning
  in convolutional neural networks, 2016.
\newblock arXiv:1604.06318.

\bibitem{li17}
Yixing Li, Zichuan Liu, Kai Xu, Hao Yu, and Fengbo Ren.
\newblock A {GPU}-outperforming {FPGA} accelerator architecture for binary
  convolutional neural networks, 2017.
\newblock arXiv:1702.06392.

\bibitem{Marcos16}
Diego Marcos, Michele Volpi, and Devis Tuia.
\newblock Learning rotation invariant convolutional filters for texture
  classification, 2016.
\newblock arXiv:1604.06720.

\bibitem{Nguyen17}
Phuc Nguyen, Ting Liu, Gautam Prasad, and Bohyung Han.
\newblock Weakly supervised action localization by sparse temporal pooling
  network, 2019.
\newblock arXiv:1712.05080.

\bibitem{Rastegari16}
Mohammad Rastegari, Vicente Ordonez, Joseph Redmon, and Ali Farhadi.
\newblock {XNOR-Net: ImageNet} classification using binary convolutional neural
  networks, 2016.
\newblock arXiv:1603.05279.

\bibitem{Redmon17}
Joseph Redmon and Ali Farhadi.
\newblock {YOLO9000: Better}, faster, stronger.
\newblock In {\em Proceedings of IEEE Conference on Computer Vision and Pattern
  Recognition (CVPR)}, July 2017.

\bibitem{Ren15}
Shaoqing Ren, Kaiming~He anbd Ross~Girshick, and Jian Sun.
\newblock {Faster R-CNN: Towards} real-time object detection with region
  proposal networks, 2015.
\newblock arXiv:1506.01497.

\bibitem{Sandler18}
Mark Sandler, Andrew Howard, Menglong Zhu, Andrey Zhmoginov, and Liang-Chieh
  Chen.
\newblock {MobileNetV2: Inverted} residuals and linear bottlenecks, 2018.
\newblock arXiv:1801.04381.

\bibitem{Sun15}
Yi Sun, Ding Liang, Xiaogang Wang, and Xiaoou Tang.
\newblock {DeepID3: Face} recognition with very deep neural networks, 2015.
\newblock arXiv:1502.00873.

\bibitem{Wai18}
Yap~June Wai, Zulkalnain bin Mohd~Yussof, Sani~Irwan bin Salim, and Lim~Kim
  Chuan.
\newblock Fixed point implementation of {Tiny-YOLO-v2} using {OpenCL} on
  {FPGA}.
\newblock {\em International Journal of Advanced Computer Science and
  Applications ({IJACSA})}, 9(10):506--512, 2018.

\bibitem{Yang16}
Shuo Yang, Ping Luo, Chen~Change Loy, and Xiaoou Tang.
\newblock {WIDER FACE: A} face detection benchmark.
\newblock In {\em {Proceedings of IEEE Conference on Computer Vision and
  Pattern Recognition (CVPR) Workshops}}, June 2016.

\bibitem{Zhang16}
Kaipeng Zhang, Zhanpeng Zhang, Zhifeng Li, and Yu Qiao.
\newblock Joint face detection and alignment using multi-task cascaded
  convolutional networks, 2016.
\newblock arXiv:1604.02878.

\bibitem{zhao17}
Ritchie Zhao, Weinan Song, Wentao Zhang, Tianwei Xing, Jeng-Hau Lin, Mani
  Srivastava, Rajesh Gupta, and Zhiru Zhang1.
\newblock Accelerating binarized convolutional neural networks with
  software-programmable {FPGA}s.
\newblock In {\em Proceedings of ACM/SIGDA International Symposium on
  Field-Programmable Gate Arrays}, pages 15--24, Feb. 2017.

\bibitem{Zhou16}
Shuchang Zhou, Yuxin Wu, Zekun Ni, Xinyu Zhou, He Wen, and Yuheng Zou.
\newblock {DoReFa-Net: Training} low bitwidth convolutional neural networks
  with low bitwidth gradients, 2016.
\newblock arXiv:1606.06160.

\end{thebibliography}
}

\end{document}